# Statistical Inference, Learning and Models in Big Data

Franke, Beate; Plante, Jean–François; Roscher, Ribana; Lee, Annie; Smyth, Cathal; Hatefi, Armin; Chen, Fuqi; Gil, Einat; Schwing, Alexander; Selvitella, Alessandro; Hoffman, Michael M.; Grosse, Roger; Hendricks, Dietrich and Reid, Nancy.

## Abstract
The need for new methods to deal with big data is a common theme in most scientific fields, although its definition tends to vary with the context. Statistical ideas are an essential part of this, and as a partial response, a thematic program on statistical inference, learning, and models in big data was held in 2015 in Canada, under the general direction of the Canadian Statistical Sciences Institute, with major funding from, and most activities located at, the Fields Institute for Research in Mathematical Sciences. This paper gives an overview of the topics covered, describing challenges and strategies that seem common to many different areas of application, and including some examples of applications to make these challenges and strategies more concrete.

## 1. Introduction

Big data provides big opportunities for statistical inference, but perhaps even bigger challenges, especially when compared to the analysis of carefully collected, usually smaller, sets of data. From January to June 2015, the Canadian Statistical Sciences Institute organized a thematic program on Statistical Inference, Learning and Models in Big Data. It became apparent within the first two weeks of the program that a number of common issues arose in quite different practical settings. This paper arose from an attempt to distill these common themes from the presentations and discussions that took place during the thematic program.

Scientifically, the program emphasized the roles of statistics, computer science, and mathematics in obtaining scientific insight from big data. Two complementary strands were introduced: cross–cutting, or foundational, research that underpins analysis, and domain–specific research focused on particular application areas. The former category included machine learning, statistical inference, optimization, network analysis, and visualization. Topic–specific workshops addressed problems in health policy, social policy, environmental science, cyber–security and social networks. These divisions are not rigid, of course, as foundational and application areas are part of a feedback cycle in which each inspires developments in the other. Some very important application areas where big data is fundamental were not able to be the subject of focussed workshops, but many of these applications did feature in individual presentations. The program started with an opening conference[1] that gave an overview of the topics of the 6–month program[2]. All the talks presented at the Fields Institute are available online through FieldsLive[3].

Technological advances enable us to collect more and more data – already in 2012 it was estimated that data collection was growing at 50% per year (Lohr, 2012). While there is much important research underway in improving the processing, storing and rapid accessing of records (Chen et al., 2014; Schadt et al., 2010), this program emphasized the challenges for modelling, inference and analysis.

Two striking features of the presentations during the program were the breadth of topics, and the remarkable commonalities that emerged across this broad range. We attempt here to illustrate many of the common issues and solutions that arose, and provide a picture of the status of big data research in the statistical sciences. While the report is largely inspired by the series of talks at the opening conference and related references, it also touches on issues that arose throughout the program.

In the next section we identify a number of challenges that are commonly identified in the world of big data, and in Section 3 we describe some examples of general inference strategies that are being developed to address these challenges. Section 4 highlights several applications, to make some of these ideas more concrete.

## 2. "Big Data, it's not the data"

There is no unique definition of big data. The title of this section is taken from the opening presentation by Bell (2015), who paraphrased Supreme Court Justice Porter Stewart by saying "I shall not today further attempt to define big data, but I know it when I see it". Bell also pointed out that the definition of 'big' is evolving with computational resources. Gaffield (2015) emphasized that people and companies are often captivated by the concept of big data because it is about us, humans. Big data provides us with detailed information – often in real time – making it possible to measure properties and patterns of human behaviour that formerly could not be measured. An analogy suggested by Keller (2015) is to the Hubble telescope: we can now see constellations with great complexity that previously appeared only as bright dots.

Beyond natural human curiosity, big data draws interest because it is valuable. A large part of the hype around big data comes from rapidly increasing investment from businesses seeking a competitive advantage (e.g. McAffee and Brynjolfsonn, 2012; Bell, 2013; Kiron and Shockley, 2011). National Statistics Agencies are exploring how to integrate big data to benchmark, complement, or create data products (Capps and Wright, 2013; Struijs et al., 2014; Letouzé and Jütting, 2014; Tam and Clarke, 2015). In many fields of research costs of data acquisition are now lower, sometimes much lower, that the costs of data analysis. The difficulty of transforming big data into knowledge is related to its complexity, the essence of which is broadly captured by the "Four Vs": volume, velocity, variety and veracity.

---

[1] The program for the opening conference is at http://www.fields.utoronto.ca/programs/scientific/14–15/bigdata/boot/
[2] A description of the complete thematic program is at https://www.fields.utoronto.ca/programs/scientific/14–15/bigdata/
[3] The Fields Institute video archive link is http://www.fields.utoronto.ca/video–archive

## 2.1 Volume

Handling massive datasets may require a special infrastructure such as a distributed Hadoop cluster (e.g. Shvachko et al., 2010) where a very large number of records is stored in separate chunks scattered over many nodes. As the costs of processors, memory and disk space continues to decrease, the limiting piece now is often communication between machines (Scott et al. 2013). As a consequence, it is impossible or impractical to access the whole dataset on a given processor and many methods will fail because they were not built to be implemented from separate pieces of data. Developments are required, and being made, to make widely–used statistical methods available for such distributed data. Two examples are the development of estimating equation theory in Lin and Xi (2011), and the consensus Monte Carlo algorithm of Scott et al. (2013).

Exploratory data analysis is a very important step of data preparation and cleaning that becomes significantly more challenging with increasing size of the data. Beyond the technical complexity of managing the data, graphical and numerical summaries are essentially low dimensional representations – for example, graphical displays are physically limited to a few million pixels. Further, the human brain requires a very low dimensional input to proceed to interpretation (Scott, 2015). One consequence of this limitation is that we often must compromise between the intuitiveness of a visualization and the power of a visualization (Carpendale et al., 2004; Nacenta et al., 2012). Another consequence is that we typically visualize only a small subset of the data at one time, and this may give a biased view (Kolaczyk, 2009; p. 91).

## 2.2 Variety

For many big data problems, the variety of data types is closely related to an increase in complexity. Much of classical statistical modelling assumes a "tall and skinny" data structure with many more rows, indexing observational or experimental units $n$, than columns, indexing variables $p$. Images, both two– and three–dimensional, blog posts, twitter feeds, social networks, audio recordings, sensor networks, videos are just some examples of new types of data that don't fit this "rectangular" model. Often rectangular data is created somewhat artificially: for example for two–dimensional images, by ordering the pixels row by row and recording one or more values associated with each pixel, hence losing information about the structure of the data.

For many types of big data the number of variables is larger, often much larger, than the number of observations. This change in the relationship between $n$ and $p$ can lead to counter–intuitive behaviour, or a complete breakdown, of conventional methods of analysis. Problems in which $p$ scales with $n$ are very common: for example, in models for networks that assign one parameter to each node, there are $p = n$ parameters for a graph with $n \times n$ possible edges (Chung and Lu, 2002). Similarly, Bioucas–Dias et al. (2013) discuss hyperspectral remote sensing image data, which may have hundreds or thousands of spectral bands, $p$, but a very small number of training samples, $n$. In a recent genomics study reported by the 1000 Genomes Project Consortium (2012), geneticists acquired data on $p = 38$ million single nucleotide polymorphisms in only $n = 1092$ individuals.

The asymptotic properties of classical methods in a regime where $p$ increases with $n$ are quite different than in the classical regime of fixed $p$. For example, if $X_i, i = 1, \ldots, n$ are independent and identically distributed as $N(0, I_p)$, and $p/n$ stays constant as $p$ and $n$ increase, the eigenvalues of the empirical covariance matrix $S = XX'/n$ do not converge to the true value of 1, but rather have a limiting distribution described by the Marcenko–Pastur law. Donoho and Gavish (2014) show how this law is relevant for understanding the asymptotic minimax risk in estimating a low rank matrix from noisy measurements by thresholding the singular values. Donoho et al. (2013) obtain similar results for the problem of matrix recovery.

## 2.3 Veracity

Some big data is obtained by piecing together several datasets that have not been collected with the current research question in mind: this is sometimes referred to as "found data" or "convenience samples". This type of data will typically have extremely heterogeneous structure (Hodas and Lerman 2014), and may well not be at all representative of the population of interest. The vast majority of big data is observational, but more importantly may be collected very haphazardly. Classical ideas of experimental design, survey sampling, and design of observational studies often get lost in the excitement of the availability of large amounts of data. Issues of sampling bias, if ignored, can lead to incorrect inference just as easily with big data as with small. Harford (2014) presents a number of high–profile failures, from the *Literary Digest* poll of 1936, which had two million participants, to Google Flu Trends in 2010, which in basing its predictions on internet search terms severely over–estimated the prevalence of influenza. Despite those limitations, National Statistical Agencies are developing methods for the careful use of such data, for example as an aid to imputation on small areas (Marchetti et al., 2015).

Cleaning large volumes of data is challenging and veracity is affected by incomplete or incorrect cleaning, and by the lack of tools to evaluate data quality. For example, to confirm that a measure of sentiment calculated from the social media was sound, Daas et al. (2015) evaluated its consistency with a well–validated indicator of consumer confidence. While it is clear that social media data are noisy and volatile, data collected from sensors may also come with several problems, including a large proportion of missing values. Daas et al. (2015) considered as an example the issues involved in cleaning traffic loop detector data.

Quality of data in large administrative databases is a pressing problem: additional heterogeneity may be introduced when the data are recorded by numerous stakeholders with varying priorities and incentives. Lix et al. (2012) use modelling to evaluate and improve the quality of the data in large health services databases.

## 2.4 Velocity

In many areas of application, data is "big" because it is arriving at a high velocity, from continuously operating instrumentation, e.g., when considering autonomous driving scenarios (Geiger et al. 2012), or through an on–line streaming service. Sequential or on–line methods of analysis may be needed, which requires methods of inference that can be computed quickly, that repeatedly adapt to sequentially added data, and that are not mislead by the sudden appearance of irrelevant data, as discussed, e.g., by Jain et al. (2006) and Hoens et al. (2012). The areas of applications using sequential learning are diverse. For instance, Li and Fei–Fei (2010) use object recognition techniques in an efficient sequential way to learn object category models from any arbitrarily large number of images from the web. Monteleoni et al. (2011) use sequential learning for climate change modelling to estimate transition probabilities between different climate models. For some applications, such as high energy physics, bursts of data arrive so rapidly that it is infeasible to even record all of it. In this case a great deal of scientific expertise is needed to ensure that the data retained is relevant to the problem of interest.

**2.5 Beyond the Vs**

Big data is about humans, it offers new possibilities, and hence, it raises new ethical questions. Duhigg (2012), for instance, reported how the Target retail stores performed analytics to send targeted marketing material to their customers, including promotions related to pregnancy to a teenager whose father was not yet aware of the coming baby. At Facebook, Kramer et al. (2014) conducted a large scale experiment on sentiments without obtaining a specific user agreement (see also Meyer, 2014).

Privacy is also a big concern, especially when considering that the linking of databases can disclose information that was meant to remain anonymous. The lawsuit following the Netflix Challenge is a striking example of that where linking the provided data to the IMDB movie reviews allowed to identify some users (Singel, 2009; 2010). In general, if released data can be linked to publicly available data, such as media articles, then privacy can be completely compromised. Another high–profile example of this was the publication of the health records of the Governor of Massachusetts, which in turn led to strict laws on the collection and release of health data in the United States known as the "HIPAA rules" (Tarran, 2014). In addition to privacy, legal questions are also unanswered: copyright and ownership of the data, for instance, is not always clear (see e.g. Struijs et al., 2014).

Sub–sampling from a large dataset is often presented as a solution to some of the problems raised in the analysis of big data; one example is the "bag of little bootstraps" of Kleiner, et al. (2014). While this strategy can be useful to analyse a dataset that is very massive, but otherwise regular, it cannot address many of the challenges that were presented in this section.

# 3. Strategies for big data analysis

In spite of the wide variety of types of big data problems, there is a common set of strategies that seem essential. Aside from pre–processing and multi–disciplinarity, many of these follow the general approach of identifying and exploiting a hidden structure within the data that often is of lower dimension. We describe some of these commonalities in this section.

**3.1 Data Wrangling**

A very large hurdle in the analysis of data, which is especially difficult for massive data sets, is manipulating the data for use in analysis. This is variously referred to as data wrangling, data carpentry, or data cleaning[4]. Big data is often very "raw": considerable pre–processing such as extracting, parsing, and storing, is required before it can be considered in an analysis.

Volume poses a challenge to data cleaning: while manual or interactive editing of a dataset are usually preferred for reasonably sized data, it quickly becomes impractical when the data grows. Large administrative databases will require special tools (e.g. the tableplots of Puts et al., 2015) to assist in the cleaning. As the size or the velocity increase, automatized solutions may become necessary. Puts et al. (2015), for instance, use a Markov model to automatize the cleaning of traffic loop data.

---

[4] The R packages dplyr and tidyr were expressly created to simplify this step, and are highly recommended. The new "pipe" operator %>%, operating similarly to the old Unix pipe command, is especially helpful. (Wickham, 2014).

Variety makes often makes data wrangling difficult, for instance when selecting variables or features from image and sound data (e.g. Guyon and Elisseeff, 2003), integrating data from heterogeneous sources (Izadi et al., 2013), or parsing free–form text for text mining or sentiment analysis (e.g., Gupta and Lehal, 2009; Pang and Lee, 2008). A more advanced approach in text mining uses the semantics of text rather than the simple count of words to describe texts (e.g., Brien et al., 2013).

One example of a data preparation problem arises in the merging of several sets of spatio–temporal data, which may have different spatial units – for example one database may be based on census tracts, another on postal codes, and still another on electoral districts. Brown et al. (2013) have developed an approach to heterogeneous complex data based on a local EM algorithm. This starts by overlaying all partitions that are used for aggregation, and then creating a finer partition such that all elements of that partitions are fully included in the larger ones. The local EM algorithm based on this is parallelizable, and hence provides a workable solution to this complex problem. A related issue arose in the presentation of Gaffield (2015), where historical census data is being used as part of a study of the impact of individuals on the history of Canada. The census tracts have changed over time, and the current database resolves this by mapping backwards from the current census tracts.

## 3.2 Visualization

Visualization is very often the first formal step of the analysis and is often a tool of choice for data cleaning (see e.g. Puts et al., 2015). The aim is to find efficient graphical representations that summarize the data and emphasise its main characteristics (Tukey, 1977). Visualization can also serve as an inferential tool in different stages of the analysis –– understanding the patterns in the data helps in creating good models (Buja et al, 2009). For example, Albert–Green et. al (2014), Woolford and Braun (2007) and Woolford et al. (2009) present exploratory data analysis tools for large environmental data sets, which are meant to not only plot the data, but also to look for structures and departures from such structures. The authors focus on convergent data sharpening and the stalagmite plot in the context of forest fire science.

Even if "a picture is worth a 1000 words", one has to realize that going from big data to 1000 word is heavy compression. Visualization is "bandwidth–limited", in the sense that there is not enough screen space to visualize large, or even moderately sized, data. Interactivity provides many ways to help with this; the most familiar is zooming into large visualizations to reveal local structure, much in the manner of Google Maps. New techniques are being developed in the information visualization community: Neumann et al. (2006) described taking cues from nature, and from fractals, in developing *PhylloTrees* for illustrating network graphs at different scales. Carpendale in her bootcamp presentation described several recent innovations in interactive visualization, many of which are illustrated at the website of the Innovis Research Group (Carpendale, 2015).

## 3.3 Reducing Dimensionality

Direct reduction of dimensionality is a very common technique: in addition to facilitating visualization (Huron et al., 2014; Wickham et al., 2010), it is often needed to address issues of data storage (Witten and Candès, 2013). Hinton and Salakhutdinov (2006) describe the use of direct dimension reduction in classification. Dimensionality reduction also often incorporates an assumption of sparsity, which is addressed in the next subsection.

Principal Component Analysis (PCA) is a classical example of dimensionality reduction, introduced by Pearson (1901). The first two, or three, principal components are often plotted in the hopes of identifying structure in the data. One of the issues with performing PCA on large data sets is the intractability of its computation, which relies on singular value decomposition (SVD). Witten and Candès (2013), Eldar and Kutyniok (2012), and Halko et al. (2011), among others, have used random matrix projections. This is a powerful tool behind compressed sensing and a striking example of dimensionality reduction, allowing approximate QR decomposition or singular value decomposition with great performance advantages over conventional methods. Witten and Candès (2013) approximate a very large matrix by the product of two "skinny" matrices that can be processed relatively easily. Using randomized methods one can infer the principal components of the original data from the singular value decomposition of these skinny factor matrices, thus overcoming a computational bottleneck.

Depending on the application, linear dimensionality reduction techniques such as PCA may need to be extended to non–linear ones, e.g., if the structure of the data cannot be captured in a linear subspace (Hinton and Salakhutdinov, 2006). However, they are harder to compute and parameterize.

In network analysis, a novel way to reduce the dimensionality of large and complex datasets is by employing an explanatory tool for data analysis called "network histogram" (Olhede and Wolfe, 2014) illustrated in Figure 1. Sofia Olhede in joint work with Patrick Wolfe suggests a non–parametric approach to estimate consistently the generating model of a network. The idea is to identify groups such that nodes from the same group show a similar connectivity pattern and to approximate the generating model by a piecewise constant function by averaging over groups. The piecewise constant function here is called stochastic block model and it approximates the generating model in the same way as a histogram approximates a pdf or the Riemann sum approximates a continuous function.

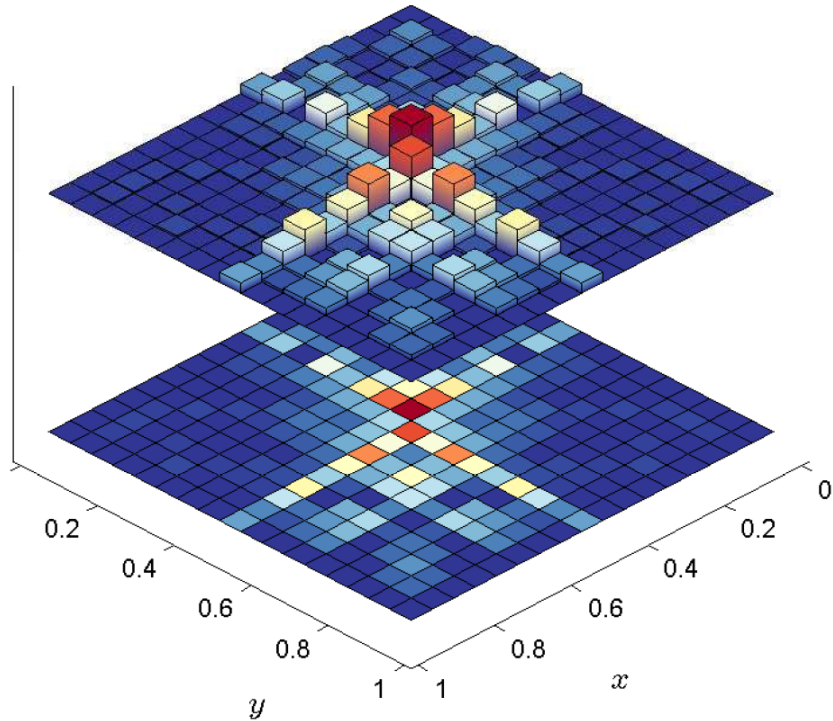

*Figure 1. Example of a network histogram from Olhede and Wolfe (2014). A network can be expressed as a matrix where each entry represent the strength of connection between two nodes. Cleverly ordering the nodes uncovers the underlying structure of the network.*

### 3.4 Sparsity and Regularization

Increasing model sparsity enforces a lower–dimensional model structure. In turn, it makesinference more tractable, models easier to interpret and it leads to more robustness against noise (Candès et al. 2008, Candès et al. 2006). Regularization techniques that enforce sparsity have been widely studied in the literature for more than a decade, including the lasso (Tibshirani, 1996), adaptive lasso (Zhou, 2006), the smoothly clipped absolute deviation (SCAD) penalty (Fan and Li, 2001), and modified Cholesky decomposition (Wu and Pourahmadi, 2003). These methods and their more recent extensions are now key tools in dealing with big data.

A strong assumption of sparsity makes it possible to use a higher number of parameters than observations, enabling us to find the optimal representation of the data while keeping the problem solvable. A good example in this context is the winning entry for the Netflix Challenge recommender system, which combined a large number of methods, thereby artificially increasing the number of parameters to improve prediction performance (Bell et al. 2008).

Sparsity–enforcing algorithms combine variable selection and model fitting: these methods typically shrink small estimated coefficients to zero and this typically introduces bias which must be taken into account in evaluating the statistical significance of the non–zero estimated coefficients. Much of the research in lasso and related methods has focussed on point estimation, but inferential aspects such as hypothesis testing and construction of confidence intervals is an active area of research. For inference about coefficients estimated using the lasso, van de Geer et al. (2014), and Javanmard and Montanari (2014a, 2014b) introduce an additional step in the estimation procedure for debiasing. Lockhart et al. (2014) in contrast focus on testing whether the lasso model contains all truly active variables. In a survival analysis setting, Fang et al. (2014) test the effect of a low dimensional component in a high dimensional setting by treating all remaining parameters as nuisance parameters. They address the problem of the bias with the aim of improving the performance of the the classical likelihood–based statistics: the Wald statistic, the score statistics and the likelihood ratio statistic, all of these based on the usual partial likelihood for survival data.

To offer a compromise to the common sparsity assumption of zero and large non–zero valued elements, Ahmed (2014) proposes a relaxed assumption of sparsity by adding weak signals back into consideration. His new high–dimensional shrinkage method improves the prediction performance of a selected submodel significantly.

### 3.5 Optimization

Optimization plays a central role in statistical inference, and regularization methods that combine model fitting with model selection can lead to very difficult optimization problems. Complex models with many hidden layers, such as arise in the construction of neural networks and deep Boltzmann machines, typically lead to non–convex optimization problems.

Sometimes a convex relaxation of an initially non–convex problem is possible: the lasso for example is a relaxation of the problem that penalizes the number of non–zero coefficient estimates, i.e. it replaces an $L_0$ penalty with an $L_1$ penalty. While this convex relaxation makes the optimization problem much easier, it induces bias in the estimates of the parameters, as mentioned above.

Non–convex penalties can reduce such bias, but non–convex optimization is usually considered to be intractable. However, statistical problems are often sufficiently well–behaved that non–convex optimization methods become tractable. Moreover, due to the structures of the statistical models, Loh and Wainwright (2014, 2015) suggest that these models do not tend to be adversarial, hence it is not necessary to look at the worst case scenario. Even when non–convex optimization algorithms, such as the Expectation–Maximization algorithm, do not provide the global optimum, the local optimum can be adequate since the distance between the local and global maxima is often smaller than the statistical error (Balakrishnan et al., 2015, Loh and Wainwright, 2015). Wainwright concluded his presentation at the opening conference with a "speculative meta–theorem", that for any problem in which a convex relaxation performs well, there is a first order simple method on the non–convex formulation which performs as good.

Applications of non–convex techniques include spectral analysis of tensors, analogous to SVD of matrices, which can be employed to aid in latent variable learning (Anandkumar et al., 2014). Non–convex methods can also be used in dictionary learning as well as in a version of robust PCA, which is less susceptible to sparse corruptions (Netrapalli et al., 2014). In some cases, convex reformulations have also been developed to find optimal solutions in non–convex scenarios (Guo and Schuurmans, 2007).

Many models used for big data are based on probability distributions, such as Gibbs distributions, for which the probability density function is only known up to its normalizing constant, sometimes called the partition function. Markov chain Monte Carlo methods attempt to overcome this by sampling from an unnormalized probability density function based on a Markov chain that has the target distribution as its equilibrium distribution, but these methods do not necessarily scale well. In some cases, approximate Bayesian computation (ABC) methods provide an alternative approximate, but computable, solution to a massive problem (Blum and Tran, 2010).

## 3.6 Measuring distance

In high–dimensional data analysis, distances between "points", or observations, are needed for most analytic procedures. For example optimization methods typically move through a possibly complex parameter space, and some notion of step–size for this movement is needed. Simulation methods such as Markov chain Monte Carlo sampling also need to 'move' through the parameter space. In both these contexts the use of Fisher information as a metric in the parameter space has emerged as a useful tool – it is the Riemannian metric (Rao, 1945; Amari and Nagaoka, 2000) when regarding the parameter space as a manifold.

Girolami and Calderhead (2011) exploited this in Markov chain Monte Carlo algorithms, with impressive improvements in speed of convergence. Grosse and Salakhutdinov (2015) showed how optimization of multi–layer Boltzmann machines could be improved by using an approximation to the Fisher information, the full Fisher information matrix being too high–dimensional to use routinely.

A similar technique emerged in environmental science (Sampson and Guttorp, 1992), where two–dimensional geographical space is mapped to a two–dimensional deformed space in which the spatial structure becomes stationary and isotropic. Schmidt et al. (2011) describe how this approach can be exploited in a Bayesian framework, mapping to a deformed space in higher dimensions, and apply this to analyses of solar radiation and temperature data.

## 3.7 Representation Learning

Representation learning, also called feature learning or deep learning, aims to uncover hidden and complex structures in data (Bengio et. al, 2013) and frequently uses a layered architecture to uncover a hierarchical data representation. The model for this layered architecture may have higher dimensionality than the original data. In that case, one may need to combine representation learning with the dimensionality reduction techniques described above in this report.

Large, well–annotated datasets facilitate successful representation learning. But labeling data for supervised learning can consume excessive time and resources, and may even prove impossible. To solve this problem, researchers have developed techniques to uncover data structure with unsupervised techniques (Anandkumar et al., 2013; Salakhutdinov and Hinton, 2012).

## 3.8 Sequential Learning

A strategy to deal with big data characterized by a high velocity is sequential or incremental learning. This technique treats the data successively, either because the task was designed in this way or the data does not allow any other access, as it is the case for continuous data streams such as long image sequences (Kalal et al. 2012). Sequential learning is also applied if the data is completely available, but has to be processed sequentially. This occurs when the data set is too large to fit into the memory or to be processed in a practically tractable way. For example, Scott et al. (2013) introduce consensus Monte Carlo, which operates by running separate Monte Carlo algorithms on distributed machines, and then averaging individual Monte Carlo draws across machines; the result can be nearly indistinguishable from the draws that would have been obtained by running a single machine algorithm for a very long time.

## 3.9 Multi–disciplinarity

A common theme in a great many presentations on big data is the necessity of a very high level of multi–disciplinarity. Statistical scientists are well–trained in both planning of studies, and in inference under uncertainty. Computer scientists bring a sophisticated mix of computational strategies and data management expertise. Both these groups of researchers need to work closely with scientists, social scientists, and humanists in the relevant application areas to ensure that the statistical methods and computational algorithms are effective for the scientific problem of interest. For instance, Dean et al. (2012) describe an exciting collaboration in forestry, where they are working on linking statistical analysis, computational infrastructure for new instrumentation, with social agency with the goal of maximizing their impact on the management of forest fires. The synergy space between them is created in an integrative approach involving scientists from all fields. This integration is not an invention of the world of big data by any means –– statistics is an inherently applied field and collaborations have been a major focus of the discipline from its earliest days. What has changed is the speed at which new technology provides new challenges to statistical analysis, and the magnitude of the computational challenges that go along with this.

Much of the big data now available involves information about people –– this is one of the aspects of big data that contributes to its popularity. The social sciences can help to elucidate potential biases in such data, as well as help to focus on the most interesting and relevant research questions. Humanists can provide frameworks for thinking about privacy, about historical data and changes through time, and about ethical issues arising in sharing of data about, for example, individuals' health. Individuals may well be willing to share their data for research purposes, for example, but not for marketing purposes: the humanities offer ways of thinking about these issues.

# 4. Examples of Applications

The breadth of applications is nearly limitless; we provide here a selection of the application areas discussed during the opening conference, in part to make the ideas outlined in the previous section more concrete.

## 4.1 Public Health

Diet is an important factor of public health in the study of disabilities (Murray et al., 2013), but assessing people's nutritional behaviour is difficult. Nielsen collects information about all products sold by a large number of groceries and corner stores at the 3–digit postal code level, and purchase card loyalty programs have data on purchases at the household level. Using these sources of information along with a UPC database of nutritional values, Buckeridge et al. (2012) match nutrition variables to existing medical records in the province of Québec, in collaboration with the *Institut de santé publique du Québec*. Since the purchase data is temporal, Buckeridge et al. (2014) measure the decrease in sales of sugary carbonated drinks following a marketing campaign, hence assessing the effectiveness of this public health effort.

Big data is also being used for syndromic surveillance, the monitoring of the syndromes of transmittable diseases at the population level. In the province of Québec, a central system monitors the triage data from emergency rooms including free form text which needs to be treated appropriately. When the automated surveillance system issues an alert, an ad–hoc analysis is performed to decide whether immediate actions are needed. Buckeridge et al. (2006) evaluate the performance of such systems through simulation for different rates of false positive alerts.

Social media produce data whose volume and variety are often challenging. The traditional reporting of flu from physician reports generally introduces a 1–2 week time delay. Google Flu Trends (Ginsberg et. al 2009) speeds up the reporting by tabulating the use of search keywords, such as "runny nose" or "sore throat", in order to estimate the prevalence of influenza–like illnesses. Predictions can sometimes go wrong, see Butler (2013), but the gain in reaction time is significant. Using the GPS location in the metadata of Twitter posts Ramírez–Ramírez et al. (2013) introduce Simulation of Infectious Diseases (SIMID) that improve geographic and temporal locality beyond Google Flu Trends. They illustrate the use of SIMID to estimate and predict flu spread in the Peel sub–metropolitan region near Toronto.

### 4.2 Health Policy

Many examples in health policy rely on the linkage of large administrative data sets. In Ontario, the Institute for Clinical Evaluative Sciences (ICES) is a province–wide research network that collects and links Ontario's health–related data at the individual level. Similar databases exists in other provinces. Comparing data on diabetes case ascertainment from Manitoba with those of Newfoundland and Labrador, Lix et al. (2012) evaluate the incompleteness of physician billing claims and estimate to which extent the number of non–fee–for–service acts is underestimated. With the quality of the data in mind, the authors also improve disease ascertainment with model–based predictive algorithms as an alternative to of relying on possibly incorrect disease codes.

Health data can be spatio temporal and complications arise quickly when the data are heterogeneous. Postal codes and census tracts, for instance, define areas that do not match, with shapes that vary through time (see e.g. Nguyen et al., 2012). In a study about cancer in Nova Scotia, Lee et al. (2015) have access to location data for the most recent years, but older data only contain the postal codes. To handle the different granularities, they use a hierarchical model where the location of cancer cases are treated as a latent variable and higher levels of the hierarchy are the aggregation units (Li et al., 2012).

### 4.3 Law and Order

Insurgencies and riots are frequent in South America, with hundreds or thousands of occurrences in each country every year. Korkmaz et al. (2015) have a database of all Twitter messages sent from South America over a period of four years as well as a binary variable named GSR, a gold standard to determine whether an event was an insurgency or not. The occurrence of an insurgency can be predicted by the volume of tweets, the presence of some keywords therein (from a total of 634 who were initially identified as potentially useful), and an increased use of *The Onion Router (*TOR), an online service to anonymize tweets. The massive database containing all the tweets is stored on a Hadoop file system and its treatment is done with Python through mapreduce. Those tools which are not part of a typical statistics cursus are standard in the treatment of massive data sets.

### 4.4 Environmental Sciences

The world is facing global environmental challenges which are hard to model, given their inherent complexity. To estimate the temperature record since 1000, Barboza et al. (2014) use different proxies including tree rings, pollen, coral, ice cores, hence combining very heterogeneous data. From earlier work, the authors found that historical records can be much better predicted by incorporating climate forcings in the model, for instance solar irradiance, greenhouse gas concentration and volcanism.

In an example combining health and environment data, Ensor et al. (2013) analyze the location and time of cardiac arrests from 2004 to 2011 in the city of Houston, with respect to air pollution measurements taken by EPA sensors scattered across the city. Instead of considering daily averages of exposition to air pollution, this study measured the exposure of an individual in the three hours preceding a cardiac arrest and find a significant effect on the risk of an out–of–hospital cardiac arrest. In an unpublished follow–up study discussed by Sallie Keller, a synthetic model is used to simulate the activities of 4.4 million people in Houston on a given day. The individual exposure varies greatly, even within a given household, as people's exposure to pollution depends on their movements in the city.The synthetic data helped optimizing CPR trainings geographically in the city.

## 4.5 Education
In a research project to study the theory of Johnson (2003) about grade inflation, Reese et al. (2014) analyze the history of Brigham Young University student evaluations from 2002 to 2014, matching it with student records. A total of 2.9 million course evaluations from 185,000 student were used. The variables of interest are three questions answered on an ordinal scale and a free text section for which a sentiment analysis determines if the comment is positive or negative. The magnitude of the Bayesian proportional odds model fitted to the ordinal response variables is too big for MCMC methods, but the ABC approach leads to a computationally feasible solution. The data shows empirical evidence of grade inflation.

With an example based on education data, Cook (2014) illustrates how visualization may be used for inference. Based on the 2012 OECD Pisa tests results from 485,900 students in 65 countries, Cook shows a figure presenting the difference in mean of math scores and reading scores between boys and girls by country. While the boys seem to outperform the girls in math, a much greater difference is seen in favour of the girls for the reading scores. To have an idea of the significance of the differences, Cook produces similar plots where the gender labels are randomly assigned to scores within each country. The fact that the true data stands out in comparison to the random assignments indicates that the gap did not occur by chance.

## 4.6 Mobile Application Security
Conveying information of interest from big data is a challenge which may be addressed by visualization. Hosseinkhani et al. (2014) propose Papilio, a novel graphical representation of a network that displays the different permissions of Android apps. The creation of a tractable representation is made possible by taking advantage of the special structure of the data. The whole representation is still too large for a single screen, so halos (Baudisch and Rosenholtz, 2003) are used to indicate the position of off screen nodes, as illustrated in Figure 2. The radius of the halos are proportional to the distance to the node.

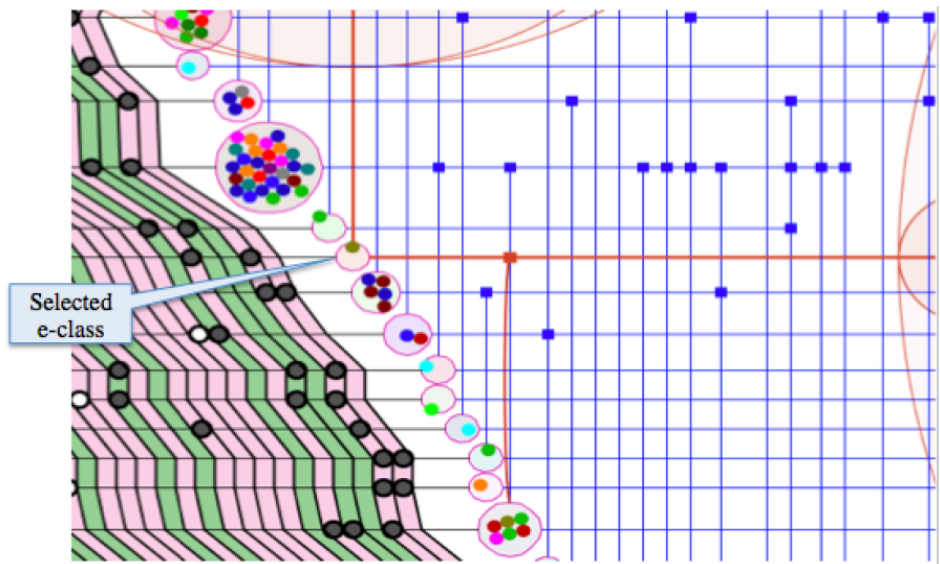

*Figure 2: Illustration of the Halos in Papilio, which indicate the position of off screen nodes. This Papilio figure from Hosseinkhani et al. (2014) displays the different permissions of Android apps.*

## 4.7 Image Recognition and Labelling

Images are unstructured data and their analysis progressed significantly with the development of deep learning models such as deep Boltzmann machines that are especially suitable for automatically learning features in the data. The Toronto Deep Learning group offer an online demo[5] of multi–modal learning based on the work of Srivastava and Salakhutdinov (2012) that can generate a caption to describe any uploaded image. The image search engine of Google also uses deep learning models to find images that are similar to an uploaded picture.

## 4.8 Digital Humanities

A large body of research in the humanities is built from an in–depth study of carefully selected authors or records. The ability to process large amounts of data is opening new opportunities where social evidence can be gathered at a population level. For instance, Drummond et al. (2006) use the 1901 Canadian census data to revise different theories about bilingualism in Canada and improve their agreement with the data. They also observe that patterns in Québec differ significantly from those in the rest of the country.

In his book *Capital in the Twenty–First Century*, Piketty (2014) discusses the evolution of references to money in novels as a reflection of inequalities in the society. Underwood et al. (2014) dig further into the question by listing all references to money amounts in English–language novels from 1750 to 1950. They find that the relative frequency of references to money amounts increases through time, but the amount mentioned decreases in constant dollar amount. They do not attempt to draw strong conclusions from this social evidence, mentioning for instance that the changes in the reading audience due to an increased literacy among the working class may also play a role in the evolution of the novels.

On a larger scale, Michel et al. (2011) analyze about 4% of all books ever printed to describe social trends through time as well as changes in linguistics. While patterns in the data match milestones events such as the Great Wars and some global epidemics, the evolution of words such as "feminism" measure important social trends. Local events such as the censorship of different artists for a number of years in different countries are also visible in the data. Those are some examples of 'culturomics', the quantitative analysis of digitized text to study cultural trends and human behavior.

---

[5] http://deeplearning.cs.toronto.edu/i2t

### 4.9 Materials Science

In an application to Materials Science by Nelson et al. (2013), the properties of a binary alloy depend on the relative position of atoms from both materials whose description relies on a very large number of explanatory variables. Surprisingly, as Candès et al. (2006) showed, a random generation of candidate materials is more efficient than a carefully designed choice. This is an example where big data shows a counter intuitive behaviour, since one could think that careful design of experiment should be able to beat random selections.

# 5. Conclusion

While technical achievements make the existence and availability of very large amounts of data possible, the challenges emerging from such data goes beyond processing, storing and accessing records rapidly. As its title indicates, the thematic program on Statistical Inference, Learning and Models for Big Data tried to focus the discussion particularly on statistical issues. There are a great many fields of research in which big data plays a key role, and it has not been possible to survey all of it.

In this report, we identify challenges and solutions that are common across different disciplines working on inference and learning for big data. With new magnitudes of dimensionality, complexity and heterogeneity in the data, researchers are faced with new challenges of scalability, quality and visualization. In addition, big data is often observational in nature, having been collected for a purpose other than the one intended, and hence making inference even more challenging.

Emerging technological solutions provide new opportunities, but they also define new paradigms for data and models. Making the models bigger with layers of latent variables may provide an insight into the features of the data, while reducing the data to lower dimensional models can be key in getting any calculations through. Lower dimensional structure analysis manifests itself not only as sparsity assumptions and dimensionality reduction, but also in visualization. Other techniques such as non–convex algorithms, sampling strategies and matrix randomization aid in the processing of large data sets. With massive data, settling for an approximate solution seems to be the saving compromise, whether by allowing for more flexibility in an objective function or by replacing a standard algorithm that is infeasible at such a large scale.

Many of the most striking examples of big data analysis presented during this thematic program came from teams of many scientists with complementary backgrounds. Moving forward, research teams need to scale up, inviting talents from different fields to come together and join forces. The phrase "Big Data" has had a great deal of media attention in recent years, creating a hype that will certainly fade. The new problems that have emerged, however, are here to stay and the need for innovative solutions will certainly keep our community busy for many years to come.

## Acknowledgments

We are grateful for the Fields Institute for Research in the Mathematical Sciences, which provided the major portion of the funding for the thematic program on Statistical Inference, Learning and Models in Big Data. The Fields Institute is funded by the Natural Sciences and Engineering Research Council of Canada and the Ministry of Training, Colleges and Universities in Ontario. Funding was also provided for specific workshops by the Pacific Institute for the Mathematical Sciences, the Centre de Recherches Mathématiques, the Atlantic Association for Advancement in the Mathematical Sciences, the Department of Statistics at Iowa State University, the National Science Foundation, and the Canadian Statistical Sciences Institute.

We thank the reviewers of an earlier version of the paper for helpful and constructive comments that improved the scope and presentation.